\documentclass[12pt,journal,compsoc]{IEEEtran}
\usepackage{cite}
\usepackage{longtable}
\ifCLASSINFOpdf
  \usepackage[pdftex]{graphicx}
\else
  \usepackage[dvips]{graphicx}
\fi
\usepackage[cmex10]{amsmath}
\usepackage{amsfonts}
\usepackage{caption}
\usepackage{subcaption}
\usepackage{multirow}
\usepackage[table,xcdraw]{xcolor}
\usepackage{booktabs}
\usepackage{tabularx}
\usepackage{listings}
\usepackage{footnote}
\usepackage{hyperref}

\usepackage{float}
\usepackage{aliascnt}
\newaliascnt{eqfloat}{equation}
\newfloat{eqfloat}{h}{eqflts}
\floatname{eqfloat}{Equation}

\newcommand*{\ORGeqfloat}{}
\let\ORGeqfloat\eqfloat
\def\eqfloat{%
  \let\ORIGINALcaption\caption
  \def\caption{%
    \addtocounter{equation}{-1}%
    \ORIGINALcaption
  }%
  \ORGeqfloat
}

\begin{document}
\title{Image Captioning}
\author{Vikram Mullachery, Vishal Motwani}
\markboth{Dec~2016}%
{Shell \MakeLowercase{\textit{et al.}}: Bare Demo of IEEEtran.cls for Computer Society Journals}
\date{\normalsize\today}

\IEEEtitleabstractindextext{%
\begin{abstract}
This paper discusses and demonstrates the outcomes from our experimentation on Image Captioning. Image captioning is a much more involved task than image recognition or classification, because of the additional challenge of recognizing the interdependence between the objects/concepts in the image and the creation of a succinct sentential narration. 
Experiments on several labeled datasets show the accuracy of the model and the fluency of the language it learns solely from image descriptions.
As a toy application, we apply image captioning to create video captions, and we advance a few hypotheses on the challenges we encountered.
\end{abstract}}

\maketitle
\IEEEpeerreviewmaketitle
\IEEEdisplaynontitleabstractindextext

\section{Introduction}
\IEEEPARstart{D}{eep} learning has powered numerous advances in computer vision tasks. One among which is Image Captioning. 
\newline
Image captioning is a much more involved task than image recognition or classification, because of the additional challenge of learning representations of the interdependence between the objects/concepts in the image and the creation of a succinct sentential narration. Here we discuss and demonstrate the outcomes from our experimentation on Image Captioning. 

Experiments on several datasets show the accuracy of the model and the fluency of the language it learns solely from image descriptions. With a handful of modifications, three of our models were able to perform better than the baseline model by A. Karpathy\footnote{\href{https://github.com/karpathy/neuraltalk2}{Neuraltalk2}}. The best BLEU and CIDEr scores that we achieved at $28.1\%$ and $0.848$ compare favorably to the baseline model’s $26.8\%$ and $0.803$, on MSCOCO dataset. 

As a toy application, we apply image captioning to create video captions, and we advance a few hypotheses on the challenges we encountered.

\section{Related Work}
Recent works in this area include Show and Tell\cite{ShowAndTell}, Show Attend and Tell\cite{ShowAttendAndTell}, among numerous others.  A highly educational work in this area was by A. Karpathy et. al. More recent advancements in this area include Review Network for caption generation by Zhilin Yang et al.\cite{ReviewNetworks} and Boosting Image Captioning with attributes by Ting Yao et al.\cite{BoostWithAttribs}.

\section{Datasets}
We use three different datasets to train and evaluate our models. These datasets contain real life images and each image in these datasets are annotated with five captions
\begin{center}
\begin{savenotes}
\begin{tabular}{ | m{6em} | m{16em} | } 
\hline
MSCOCO  \footnote{\href{http://cocodataset.org/}{MSCOCO Dataset}} \cite{mscoco}  & Contains 120K images with 5 captions for each split : 80k images for Training and 40k images for Validation  \\
\hline
Flickr30k \footnote{\href{http://web.engr.illinois.edu/~bplumme2/Flickr30kEntities/}{Flickr30k Dataset}}\cite{flickr30k} & Contains 30K images with 5 captions each split : 28K images for Training and 2k images for validation  \\
\hline
Flickr8k \footnote{\href{http://nlp.cs.illinois.edu/HockenmaierGroup/Framing_Image_Description/KCCA.html}{Flickr8k Dataset}} \cite{flickr8k} & Contains 8K images with 5 captions each split : 7k images for training and 1k images for validation    \\
\hline
\end{tabular}
\end{savenotes}
\end{center}

\section{General Architecture}
The goal is to maximize the probability of the correct description given the image by using the following formalism: \\
\begin{eqfloat}
\begin{equation}
{\theta^* = {arg\,max}_{\theta} \sum_{I,S} \log P(S|I;\theta)}
\end{equation}
\caption{Objective function where $\theta$ are the parameters of the model, $I$ is an image, and $S$ its correct transcription
}
\end{eqfloat}
\\
Since $S$ represents any sentence, its length is unbounded. Thus, it is common to apply the chain rule to model the joint probability over ${S_0 ,. . . , S_N}$ where $N$ is the length of this particular sentential transcription (also called caption) as \\
\begin{eqfloat}
\begin{equation}\label{eqn:sentprob}
{logP (S|I;\theta) = \sum_{t=0}^{N} log(S_t | I, S_0, S_1, ... S_{t-1};\theta)}
\end{equation}
\caption{Modeling Sentence Probability}
\end{eqfloat}
\\
At training time, ${(S, I)}$ is a training example pair, and we optimize the sum of the log probabilities as described in equation \ref{eqn:sentprob} over the whole training set using Adam optimizer\footnote{\href{https://machinelearningmastery.com/adam-optimization-algorithm-for-deep-learning/}{Adam Optimization}}. It is natural to model ${P(S_t | I, S_0, S_1, ... S_{t-1})}$ with a Recurrent Neural Network (RNN), where the variable number of words we condition upon up to $t-1$ is expressed by a fixed length hidden state or memory $h_t$. This memory is updated after seeing a new input ${x_t}$ by using a nonlinear function ${f : h_{t+1} = f(h_t,x_t) }$ . For $f$ we use a Long-Short Term Memory (LSTM) network. For the representation of images, we use a Convolutional Neural Network (CNN). CNNs have been widely used and studied for image tasks, and is considered, currently, the state-of-art for object recognition and detection. LSTMs and other variants of RNNs have been studied extensively and used widely for time recurrent data such as words in a sentence or the next time step's stock price etc.
\\
\begin{figure}[h]
\includegraphics[width=8cm]{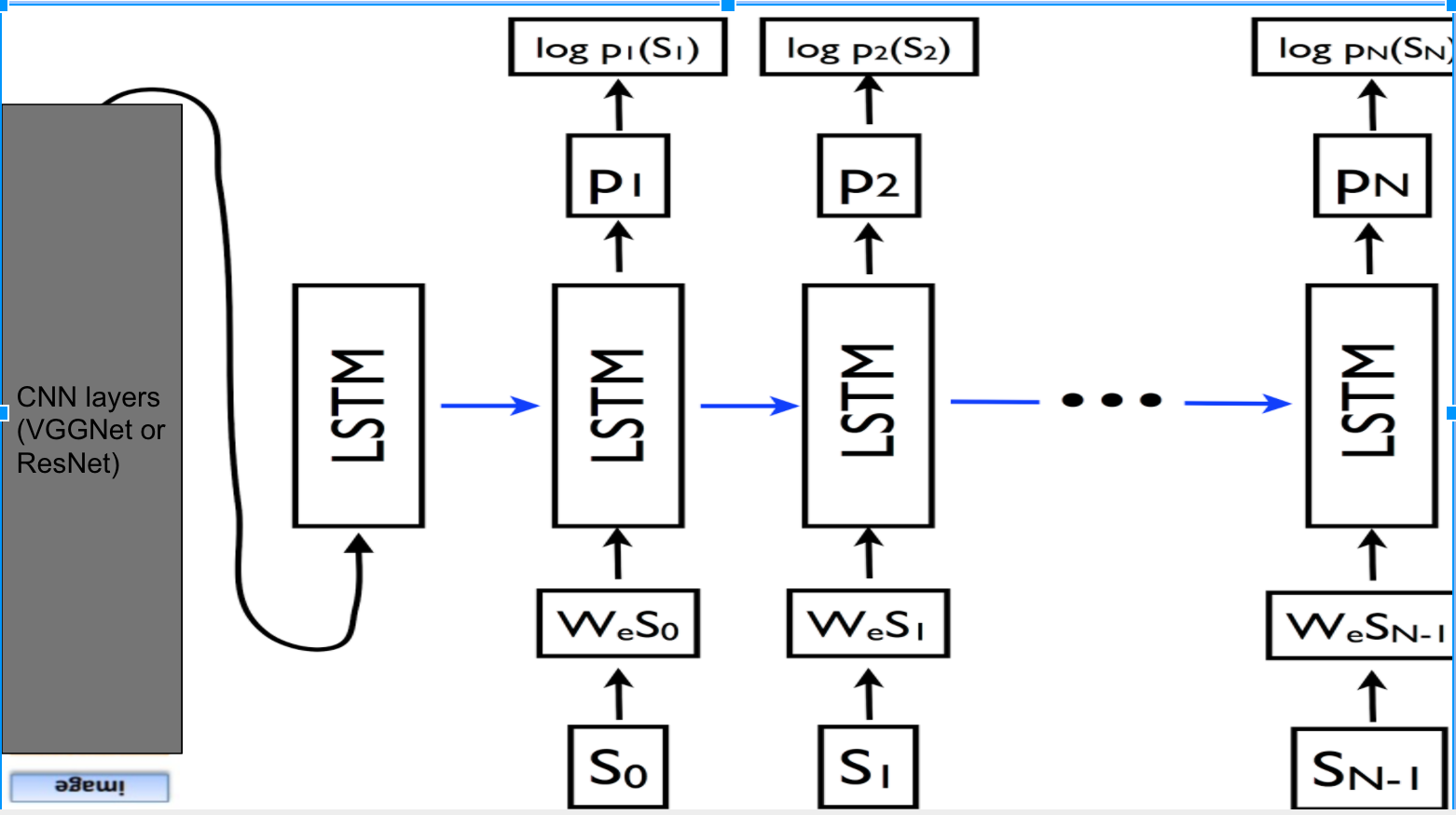}
\caption{LSTM decoder combined with CNN image encoder. The unrolled connections between the LSTM memories are in blue and they correspond to the recurrent connections. All LSTMs share the same parameters}
\centering
\end{figure}

The LSTM model is trained to predict each word of the sentence after it has seen the image as well as all preceding words as defined by ${P(S_t | I, S_0, S_1, ... S_{t-1})}$. For this purpose, it is instructive to think of the LSTM in unrolled form; a copy of the LSTM memory is created for the image and each sentence word such that all LSTMs share the same parameters and the output ${m_t - 1}$ of the LSTM at time $t - 1$  is fed to the LSTM at time $t$ (see Figure 1). All recurrent connections are transformed to feed-forward connections in the unrolled version. In more detail, if we denote by $I$ the input image and by ${S = S_0, . . . , S_N}$ a true sentence describing this image, the unrolling procedure reads \\
\begin{eqfloat}
\begin{equation}
{x_{-1} = CNN(I)}
\end{equation}
\begin{equation}
{x_{t} = W_e S_t, t \in \{0 ... N-1 \}}
\end{equation}
\begin{equation}
{p_{t+1} = LSTM(x_t), t \in \{0 ... N-1 \}}
\end{equation}
\end{eqfloat}
\\
where we represent each word as a one-hot vector $S_t$ of dimension equal to the size of the dictionary. Note that we denote by $S_0$ a special start word and by $S_N$ a special stop word which designates the start and end of the sentence. In particular, by emitting the stop word the LSTM signals that a complete sentence has been generated. Both the image and the words are mapped to the same space, the image by using a vision CNN, the words by using word embedding $W_e$. The image $I$ is only input once, at $t = -1$, to inform the LSTM about the image contents. 

We use negative log likelihood loss: \\
\begin{eqfloat}
\begin{equation}
{L(I,S) = - \sum_{t=1}^{N}{\log P_t(S_t)}}
\end{equation}
\end{eqfloat}
\\
The above loss is minimized with respect to all the parameters of the LSTM, from the top layer of the image embedder CNN to the word embedding ${W_e}$.
Further, to generate sentence, beam search is used. It iteratively considers the set of $k$ best sentences up to time $t$ as candidates to generate sentences of size ${t + 1}$, and retains only the best $k$ of them. This approximates ${S = argmax_{S'} P(S' | I)}$. We use beam size of $20$ in all our experiments.

\section{Baseline Model}
We use A. Karpathy’s pretrained model as our baseline model. This model is trained only on MSCOCO dataset. The model uses a $16$-layer VGG Net for embedding image features which is fed only to the first time step of the single layer RNN which is constituted of long-short term memory units (LSTM). The RNN size in this case is $512$. Since words are one hot encoded, the word embedding size and the vocabulary size is also $512$. 

The two parts, CNN and RNN, are joined together by an intermediate feature expander, that feeds the output from the CNN into the RNN. Recall, that there are $5$ labeled captions for each image. The feature expander allows the extracted image features to be fed in as an input to multiple captions for that image, without having to recompute the CNN output for a particular image. 

In VGG-Net, the convolutional layers are interspersed with maxpool layers and finally there are three fully connected layers and softmax. The softmax layer is required so that the VGGNet can eventually perform an image classification. But for the purpose of image captioning, we are interested in a vector representation of the image and not its classification. And so, the last two layers are eliminated and the output from the fully connected layer can be extracted and expanded to feed into the RNN part of the architecture. 

\section{Experiments and Modifications}
We attempted three different types of improvisations over the baseline model using controlled variations to the architecture.
\subsection{Transfer Learning: Flickr8k/30k}
First improvement was to perform further training of the pretrained baseline model on Flickr8K and Flickr30k datasets. After building a model identical to the baseline model \footnote{\href{http://cs.stanford.edu/people/karpathy/neuraltalk2/checkpoint_v1.zip}{Downloadable baseline model}}, we initialized the weights of our model with the weights of the baseline model and additionally trained it on Flickr 8k and Flickr 30K datasets, thus giving us two models separate from our baseline model

\subsection{RNN hidden layers}
Second improvement was increasing the number of RNN hidden layers over the baseline model. When we add more hidden layers to the RNN architecture, we can no longer start our training by initializing our model using the weights obtained from the baseline model (since it consists of just $1$ hidden layer in RNN architecture). Hence in this case we pre-initialize the weights of only the CNN architecture i.e VGGNet by using the weights obtained from deploying the same 16 layer VGGNet on an ImageNet classification task. Thus using this method, we were able to increase the number of hidden layers in the RNN architecture to two (2) and four (4) layers.

\subsection{ResNet in lieu of VGGNet}
The third improvement was to use ResNet (Residual Network)\cite{ResNet} in place of VGGNet. Our Motivation to replace VGG Net with Residual Net (ResNet) comes from the results of the annual Imagenet classification task. Following are the results for the imagenet classification task over the years
\begin{figure}[H]
\includegraphics[scale=.15]{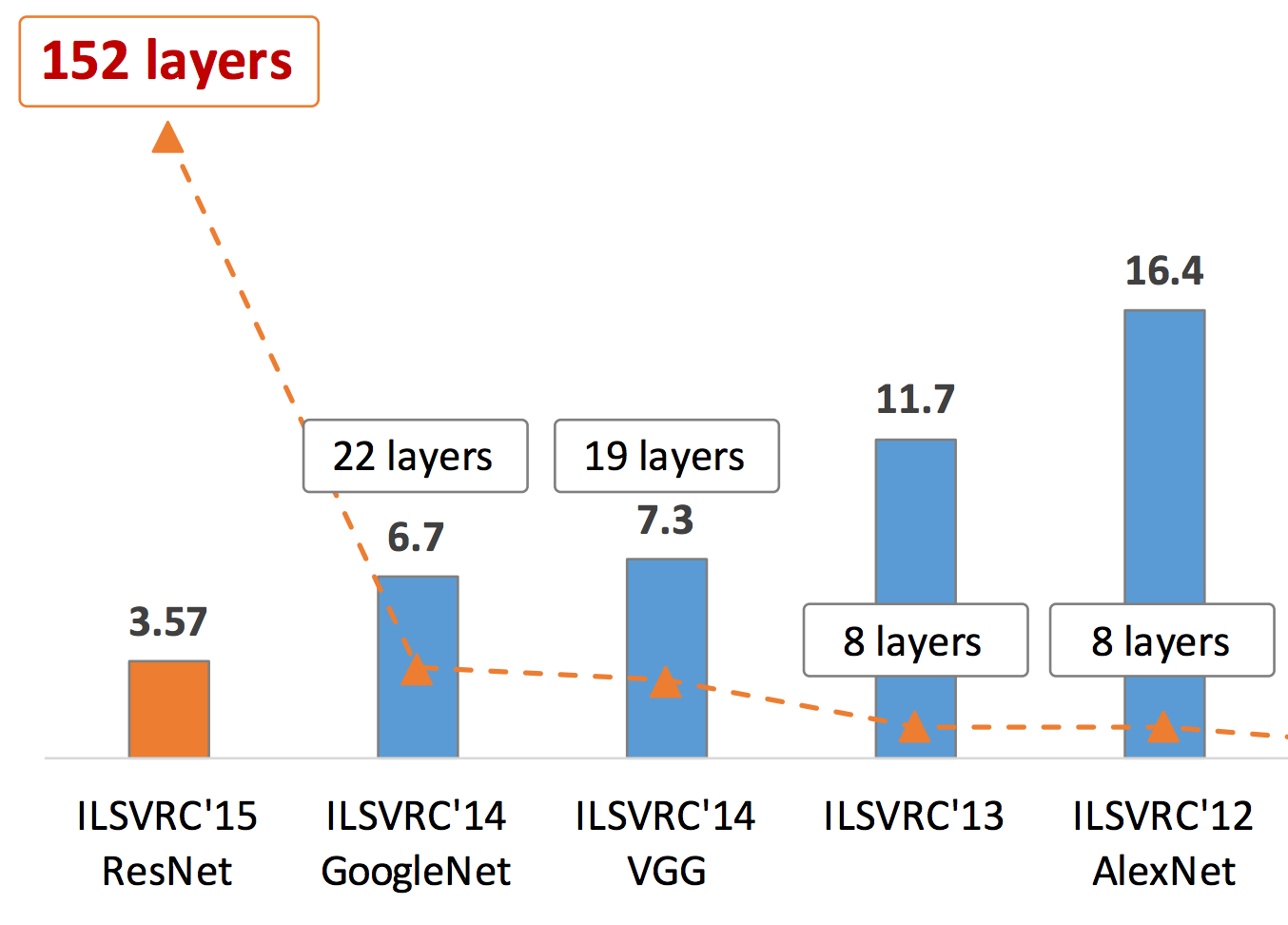}
\caption{ImageNet Results over time}
\end{figure}

It has been empirically observed from these results and numerous others, that ResNet can encode better image features.
ResNet architecture is a 100 to 200 layer deep CNN. To account for the problem of vanishing gradients, ResNet has the following scheme of skip connections.\\
\begin{figure}[H]
\includegraphics[scale=.15]{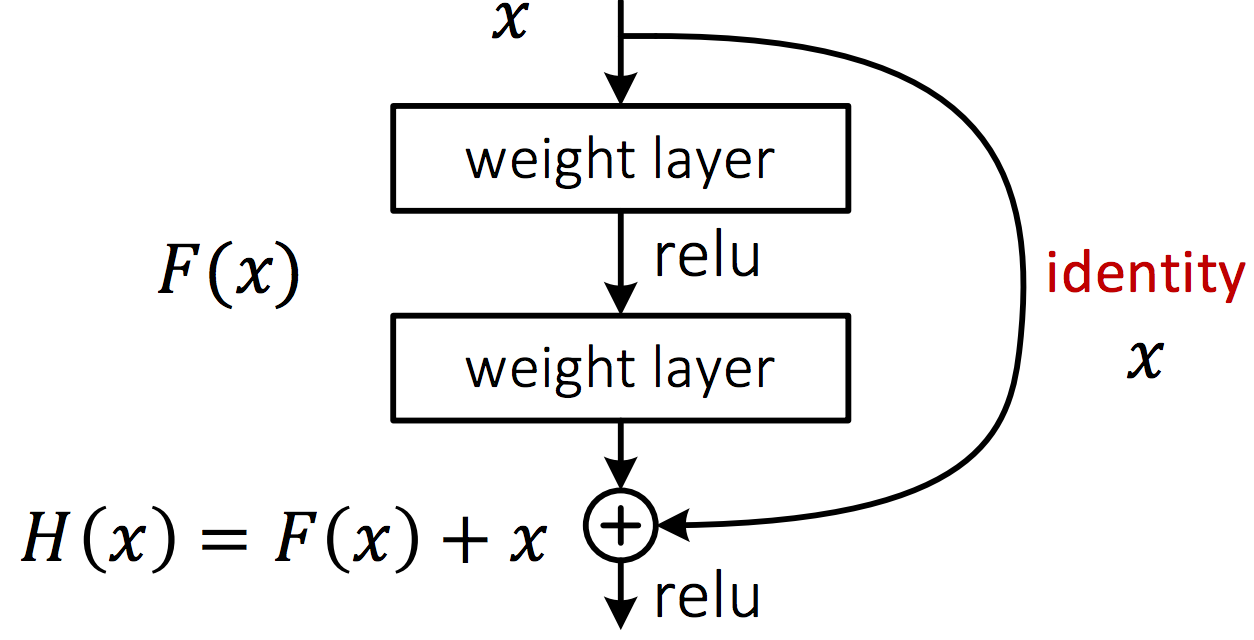}
\caption{Skip Connections in ResNet}
\end{figure}

Inspired from the results of ResNet on Image Classification task, we swap out the VGGNet in the baseline model with the hope of capturing better image embeddings.
\begin{figure}[H]
\includegraphics[scale=.3]{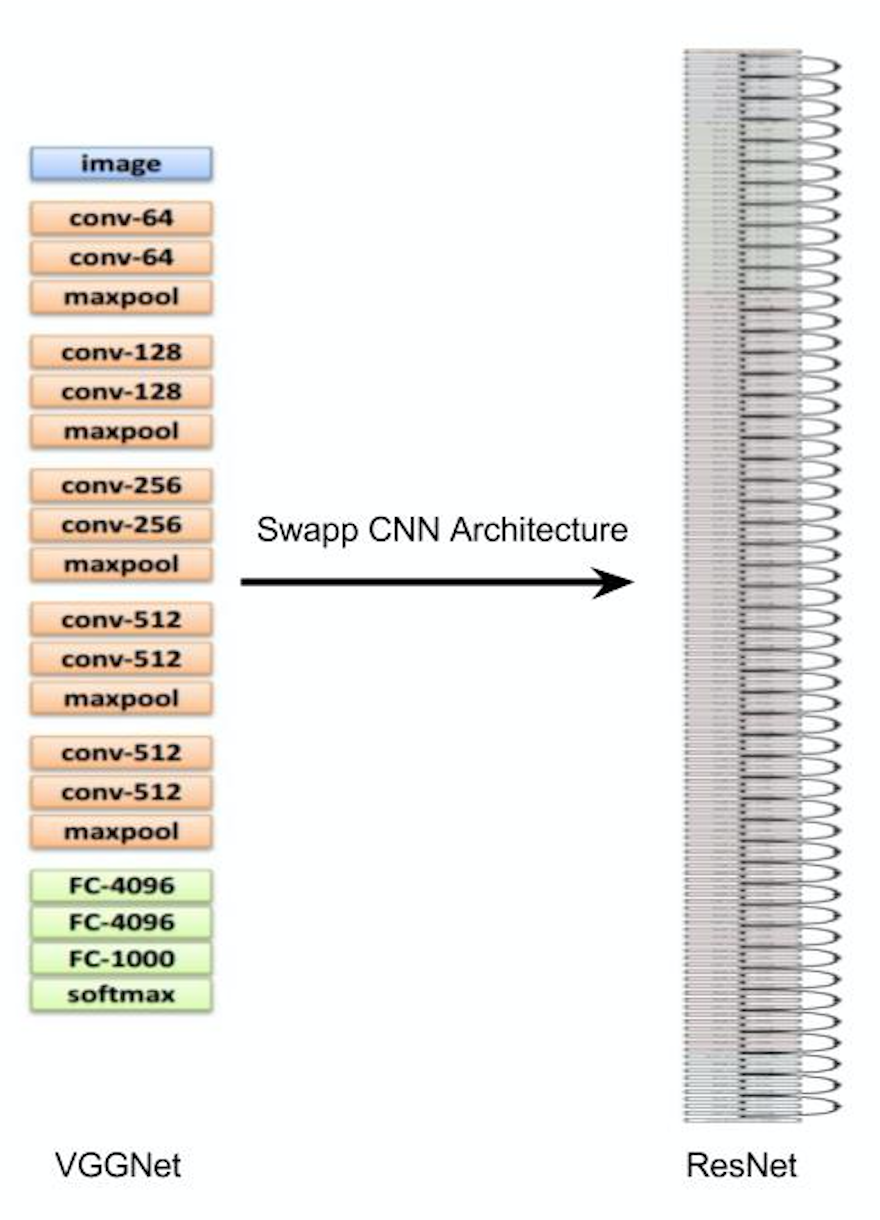}
\caption{Swap VGGNet with ResNet}
\end{figure}

We use 101 layer deep ResNet for our experiments.
\begin{figure}[H]
\includegraphics[scale=.14]{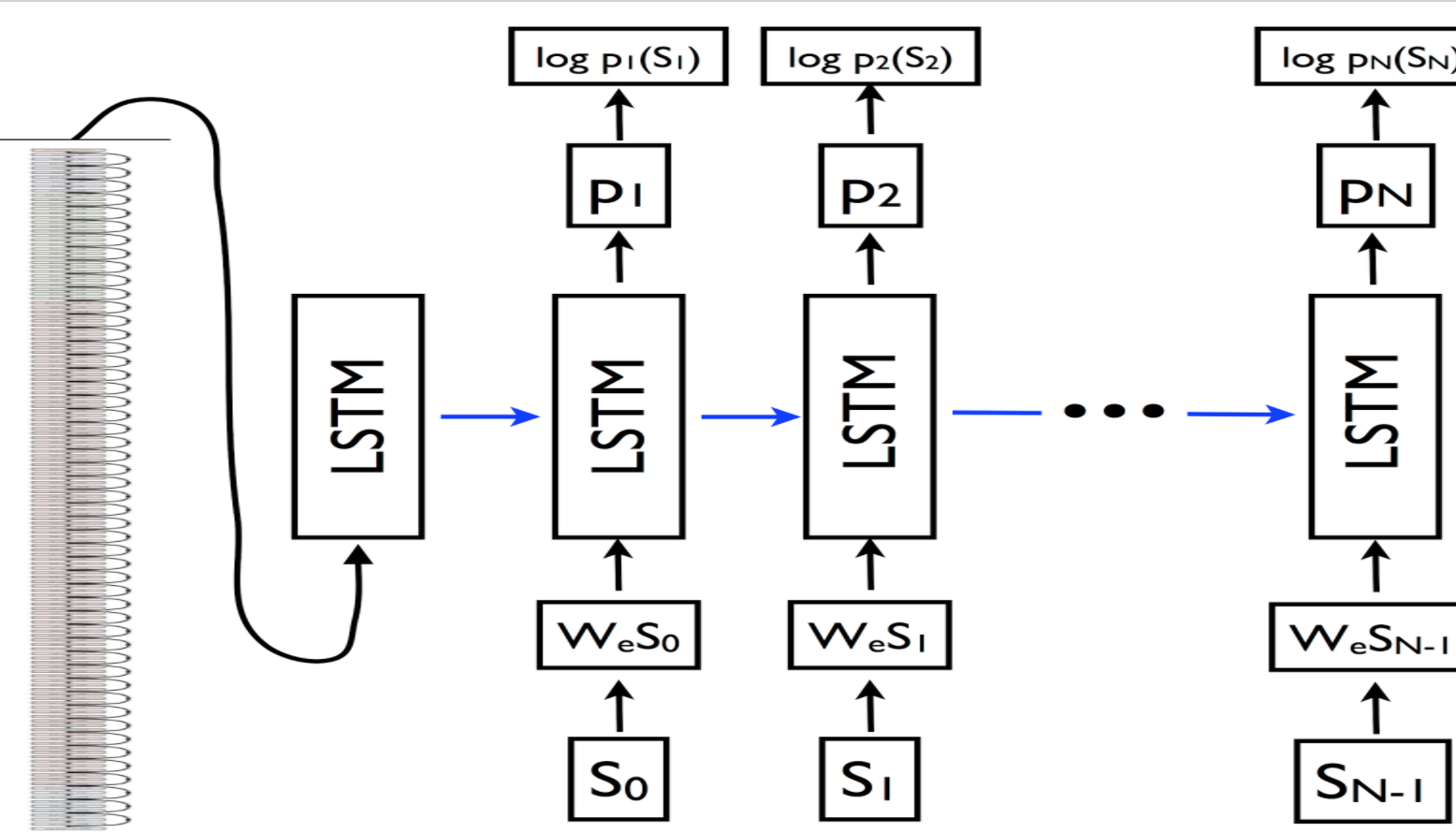}
\caption{Architecture: ResNet for Encoding}
\end{figure}

We pre initialize the weights of only the CNN architecture i.e ResNet by using the weights obtained from deploying the same ResNet on an ImageNet classification task. Note that there are no changes to the RNN portion of the architecture for this experimentation choice.

\section{Evaluation Metrics}
There are two evaluation metrics of interest to us. First, a caption language evaluation score, \texttt{BLEU\_4} \footnote{\href{https://en.wikipedia.org/wiki/BLEU }{BLEU score}} (bilingual evaluation understudy) score, which is an algorithm for evaluating the quality of text which has been machine-translated from one natural language to another. It ranges from $0$ to $1$, with $1$ being the best score, approximating a human translation.\\

Second, CIDEr\footnote{\href{https://arxiv.org/pdf/1411.5726.pdf}{CIDEr: Consensus-based Image Description Evaluation}} score, which is a consensus-based evaluation protocol for image description evaluation, which enables an objective comparison of machine generation approaches based on their “human-likeness”, without having to make arbitrary calls on weighing content, grammar, saliency, etc. with respect to each other. This score is usually expressed as a percentage or a fraction, with $100\%$ indicating human generated caption for an image. 

\section{Results and Discussion}
Following is a listing of the models that we experimented on: 
\begin{center}
\begin{savenotes}
\begin{tabular}{ | m{4em} | m{14em} | } 
\hline
Baseline  & PreTrained model - A.Karpathy’s work (Trained only on MSCOCO)  \\
\hline
Model 1 & Additional Training of Baseline on Flickr8k  \\
\hline
Model 2 & Additional Training of Baseline on Flickr30k \\
\hline
Model 3 & VGGNet 16-layer with 2 layer RNN (Trained ONLY on MSCOCO) \\
\hline
Model 4 & VGGNet 16-layer with 4 layer RNN (Trained ONLY on MSCOCO) \\
\hline
Model 5 & ResNet 101-layer with 1 layer RNN (Trained ONLY on MSCOCO) \\
\hline
\end{tabular}
\end{savenotes}
\end{center}

Following are a few key hyperparameters that we retained across various models. These could be helpful for attempting to reproduce our results. 

\begin{center}
\begin{savenotes}
\begin{tabular}{ | m{9em} | m{9em} | } 
\hline
RNN Size  & 512  \\
\hline
Batch size & 16 \\
\hline
Learning Rate & 4e-4 \\
\hline
Learning Rate Decay & 50\% every 50000 iterations \\
\hline
RNN Sequence max length & 16 \\
\hline
Dropout in RNN & 50\% \\
\hline
Gradient clip & 0.1\% \\
\hline
\end{tabular}
\end{savenotes}
\end{center}

Following are the results in terms of \texttt{BLEU\_4} scores and CIDEr scores of the various models on the different datasets.  \\
\begin{figure}[H]
\includegraphics[width=8cm]{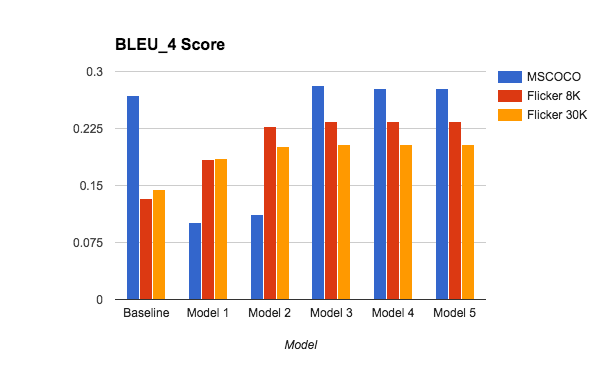}
\caption{\texttt{BLEU\_4} score}
\centering
\end{figure}

\begin{figure}[H]
\includegraphics[width=8cm]{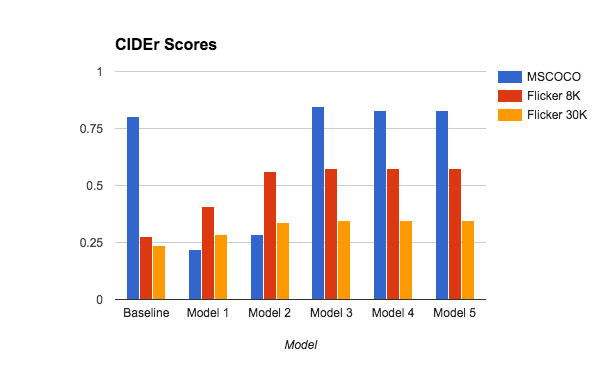}
\caption{CIDEr score}
\centering
\end{figure}

Following graph shows the drop in cross entropy loss against the training iterations for VGGNet + 2 RNN model (Model 3). Note that each iteration corresponds to one batch of input images. In our experiments, Model 3 outperformed all the other models. \\

\begin{figure}[H]
\includegraphics[width=8cm]{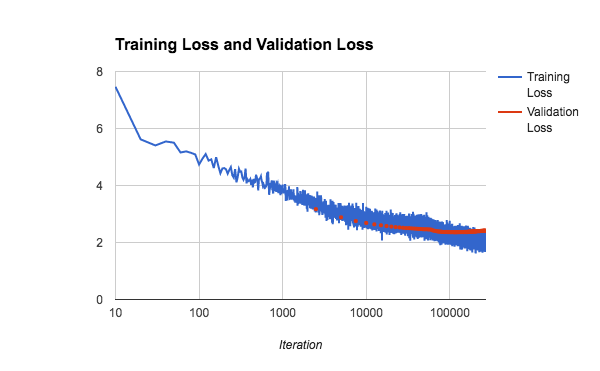}
\caption{Learning Rate for Model 3 (VGGNet with 2 layer RNN)}
\centering
\end{figure}

\textbf{Discussion of a few results} \\
A few instances of correct captions: \\
\begin{center}
\begin{savenotes}
\begin{table}[H]
\begin{tabular}{ | m{25em} | } 
\hline
\\[1em]
\begin{subfigure}{0.9\columnwidth}
\includegraphics[width=\linewidth]{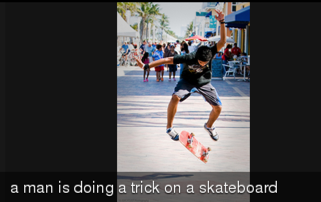}
    \caption{Note that this is not a copy of any training image caption, but a novel caption generated by the system.}
\end{subfigure}\\[1em]
\begin{subfigure}{0.9\columnwidth}
\includegraphics[width=\linewidth]{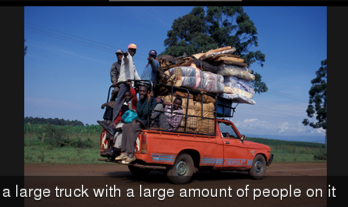}
    \caption{Similar to the above, this a novel caption, demonstrating the ability of the system to generalize and learn}
\end{subfigure} \\ [1 em]
\hline
\end{tabular}
\captionsetup{labelformat=empty}
\caption{Fig 9.   Good Captions}
\end{table}
\end{savenotes}
\end{center}

Hypotheses on incorrect captions:\\

\begin{center}
\begin{savenotes}
\begin{table}[H]
\begin{tabular}{ | m{25em} | } 
\hline
\\[1em]
\begin{subfigure}{0.9\columnwidth}
\includegraphics[width=\linewidth]{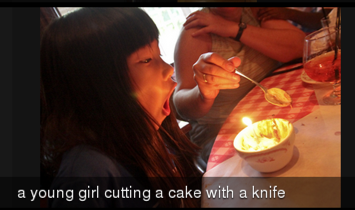}
    \caption{High co-occurrences of “cake” and “knife” in training data and zero occurrences of “cake” and “spoon”, thus engendering this caption}
\end{subfigure}\\[1em]
\begin{subfigure}{0.9\columnwidth}
\includegraphics[width=\linewidth]{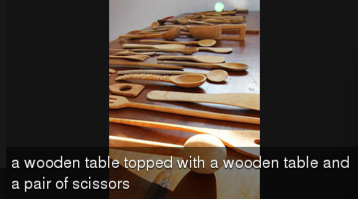}
    \caption{High occurrences of “wooden” with “table”, and then further with “scissors”. Zero occurrences of word “wooden” with the word “utensils” in training data.  Further, this caption shows vulnerability of the model in that the caption could be nonsensical to a human evaluator }
\end{subfigure} \\ [1 em]
\hline
\end{tabular}
\captionsetup{labelformat=empty}
\caption{Fig 10.  Poor Captions}
\end{table}
\end{savenotes}
\end{center}

\section{Toy Application - Video captioning}
As an experimentation to apply video captioning in real-time we loaded a saved checkpoint of our model and generated a caption of the video frame. Following are some amusing results, both agreeable captions\footnote{\href{https://youtu.be/-OqPY16WkVc}{Correct video captions}} and poor captions\footnote{\href{https://youtu.be/96Is2T\_Suqk}{Poor video captions}}.

Entertaining as some of the above maybe, they teach us a few valuable things about video captioning being different from static image captioning. Empirically, one observes that there are abrupt changes in captions from one frame to the next. However, intuitively and experientially one might assume the captions to only change slowly from one frame to another. This disconnect would suggest feeding the caption from one frame as an input to the subsequent frame during prediction. 

Additionally, the current video captioning sways widely from one caption to another with very little change in camera positioning or angle. This rapid change in caption appears to be akin to a highly sensitive decoder. This demonstrates a dearth of inertia or recognition of the image source as a video from a camera (as opposed to disconnected slides of individual images). Consequently, this  would suggest the necessity to stabilize/regularize  the caption from one frame to the next. 

A third item to watch out for is the apparent unrelated and arbitrary captions on fast camera panning. Since this is an expected real-life action on a camera, there will need to be, as yet unexplored, adjustments and accommodations made to the prediction method/model.

\section{Future Work}
We observe that ResNet is definitely capable of encoding better feature vector for images. Also, taking tips from the current state of art, i.e show attend and tell, it should be of interest to observe the results that could be obtained from applying attention mechanism on ResNet. 
For the decoder we currently do not use the dense embedding of words. Also, we do not initialize the weights of RNN architecture from the weights of a pre trained language model. Though Vinyals et al. mention that they do not observe any significant gain by pre-training the RNN language model, it should be of interest to observe if it’s the same scenario when used in conjunction with ResNet. 
Ensembles have long been known to be a very simple yet effective way to improve performance of machine learning systems. In the context of deep architectures, one only needs to train separately multiple models on the same task, potentially varying some of the training conditions, and aggregating their answers at inference time. This is another effort that should be worth pursuing in future work.

\section{Acknowledgement}
K. Simonyan and A. Zisserman. Visual Geometry Group. Very Deep Convolutional Networks for Large-Scale Visual Recognition \cite{VGGNet} \\

Tsung{-}Yi Lin and Michael Maire and Serge J. Belongie and Lubomir D. Bourdev and Ross B. Girshick and James Hays and C. Lawrence Zitnick. MSCOCO dataset\cite{mscoco}\\

Bryan A. Plummer,  Liwei Wang, Christopher M. Cervantes, Juan C. Caicedo, Julia Hockenmaier,  Svetlana Lazebnik. Flickr30k dataset. \cite{flickr30k}\\

Cyrus Rashtchian, Peter Young, Micah Hodosh, and Julia Hockenmaier. Flickr8k dataset
 \cite{flickr8k} \\

\ifCLASSOPTIONcaptionsoff
  \newpage
\fi

\bibliography{ref.bib}
\bibliographystyle{IEEEtran}
\newpage

\end{document}